\documentclass[conference,letterpaper]{IEEEtran}
\makeatletter
\def\IEEEtitletopspaceextra{19pt} 
\makeatother
\IEEEoverridecommandlockouts
\usepackage{amsmath,amssymb,amsfonts}
\usepackage{algorithmic}
\usepackage{graphicx}
\usepackage{textcomp}
\usepackage{booktabs}
\usepackage[caption=false,font=footnotesize]{subfig}

\usepackage{stfloats}
\usepackage{siunitx}
\usepackage{xcolor}
\def\BibTeX{{\rm B\kern-.05em{\sc i\kern-.025em b}\kern-.08em
    T\kern-.1667em\lower.7ex\hbox{E}\kern-.125emX}}
\graphicspath{{Images/}}
\begin{document}

\title{Humanizing Robot Gaze Shifts: A Framework for Natural Gaze Shifts in Humanoid Robots}

\author{
Jingchao Wei, Jingkai Qin, Yuxiao Cao, Jingcheng Huang, Xiangrui Zeng\textsuperscript{*}, Min Li, Zhouping Yin
\thanks{* Corresponding author.}
\thanks{Jingchao Wei, Jingkai Qin, Yuxiao Cao, Jingcheng Huang, Xiangrui Zeng, Min Li and Zhouping Yin are with the School of Mechanical Science and Engineering, Huazhong University of Science and Technology, Wuhan 430074, China (e-mail: weijingchao@hust.edu.cn; corresponding author e-mail: zeng@hust.edu.cn).}
\thanks{This work was supported by the Hubei Province Key Research and Development Project (No. 2025BEB008) and the Fundamental and Interdisciplinary Disciplines Breakthrough Plan of the Ministry of Education of China.}
}
\maketitle

\begin{abstract}

Leveraging auditory and visual feedback for attention reorientation is essential for natural gaze shifts in social interaction. However, enabling humanoid robots to perform natural and context-appropriate gaze shifts in unconstrained human--robot interaction (HRI) remains challenging, as it requires the coupling of cognitive attention mechanisms and biomimetic motion generation. In this work, we propose the Robot Gaze-Shift (RGS) framework, which integrates these two components into a unified pipeline. First, RGS employs a vision--language model (VLM)-based gaze reasoning pipeline to infer context-appropriate gaze targets from multimodal interaction cues, ensuring consistency with human gaze-orienting regularities. Second, RGS introduces a conditional Vector Quantized-Variational Autoencoder (VQ-VAE) model for eye--head coordinated gaze-shift motion generation, producing diverse and human-like gaze-shift behaviors. Experiments validate that RGS effectively replicates human-like target selection and generates realistic, diverse gaze-shift motions.

\end{abstract}


\section{Introduction}

With the rapid advancement of robotics and artificial intelligence technologies, human--robot interaction (HRI) has emerged as a pivotal research topic in robotics, particularly in social robotics. The core objective of this research field is to develop robots capable of natural, intuitive, and meaningful interactions with humans \cite{vsabanovic2023robots}. In recent years, significant progress has been made in both verbal and nonverbal interaction, such as dialogue management \cite{jokinen2025domain}, limb motion generation \cite{he2024omnih2o}, and facial expression generation \cite{liu2024unlocking}. However, specialized studies focusing on \emph{robot gaze shift}---i.e., how robots dynamically reorient their gaze targets over time---remain relatively scarce. Empirical studies in social and cognitive science suggest that gaze shifts play a fundamental role in coordinating attention, aligning cognitive processes, and establishing socio-emotional connections among individuals \cite{bristow2007social}, \cite{frischen2007gaze}. Similarly, robotics studies have shown that timely, appropriately directed gaze shifts by robots can increase user engagement and foster emotional rapport \cite{admoni2017social}. Therefore, endowing robots with human-like gaze-shift behaviors is a promising way to improve user experience in HRI. Achieving this requires two complementary capabilities: (i) gaze reasoning, i.e., inferring gaze targets from interaction context by following human perceptual and behavioral regularities; (ii) gaze-shift motion generation, i.e., synthesizing natural eye--head coordinated movements that resemble human gaze shifts.

For robots to achieve anthropomorphic gaze reasoning, they must ground gaze-target selection in the regularities that govern human attention and social perception, including the orienting reflex \cite{zhou2023orienting} and responsive joint attention \cite{mundy2007attention}. Importantly, many of these principles can be described as clear and intuitive natural-language rules; for instance, people almost invariably shift their gaze toward someone who waves at them. Yet translating these high-level principles into a robotic system and integrating the necessary perception, context understanding, and decision-making modules into a robust pipeline remains a substantial challenge.

Complementary to gaze reasoning, realizing the second capability---gaze-shift motion generation---requires robots to produce human-like eye--head coordination when reorienting gaze. Under fixed-torso conditions, human gaze reorientation relies on tightly coupled eye and head movements to support a suite of gaze behaviors. However, a major technical bottleneck arises from the pronounced inter-individual variability in human gaze motor strategies. For example, some individuals adopt eye-dominant strategies when executing gaze shifts, whereas others rely more heavily on head-dominant adjustments. Such heterogeneity makes it difficult to formulate a single, one-size-fits-all quantitative model that faithfully captures the regularities of human gaze-shift kinematics.

In this paper, we propose a novel Robot Gaze-Shift (RGS) framework for human-like gaze reasoning and gaze-shift motion generation in unconstrained HRI. Gaze shifts constitute a key nonverbal mechanism in human communication, conveying subtle socio-emotional cues and signaling attention, engagement, and active listening. By endowing robots with human-like gaze-shift behaviors, our goal is to make HRI more natural and engaging. The proposed RGS framework integrates gaze reasoning and gaze-shift motion generation into a unified pipeline, enabling robots to infer context-appropriate gaze targets and to realize them through coordinated eye--head movements. The main contributions are summarized as follows:

\begin{itemize}
    \item We propose a vision--language model (VLM)-based gaze reasoning pipeline that enables robots to infer context-appropriate gaze targets from multimodal interaction cues in unconstrained HRI.
    \item We introduce a conditional Vector Quantized-Variational Autoencoder (VQ-VAE) model for eye--head coordinated gaze-shift motion generation, enabling robots to produce diverse and human-like gaze-shift behaviors under varying interaction scenarios.
    \item Experiments on scenario-based video stimuli, our self-collected human gaze-shift dataset, and a desktop humanoid robot head validate the proposed RGS framework: the VLM-based gaze reasoning pipeline selects gaze targets consistent with human gaze-orienting regularities, while the conditional VQ-VAE model generates naturalistic and diverse eye--head coordinated gaze shifts.
\end{itemize}

\section{Related Work}

\subsection{Robot Gaze Reasoning}

Prior work on robot gaze reasoning can be broadly grouped into two categories: heuristic approaches and data-driven methods. Specifically, heuristic approaches employ hand-crafted rules inspired by findings from human--human interaction to regulate robotic gaze-target selection \cite{zhang2017look}, \cite{pan2020realistic}, \cite{mishra2022knowing}, \cite{correia2023robotic}. In contrast, data-driven methods leverage human interaction data to learn models that directly predict robots' gaze targets from observed cues \cite{vogel2008targetdirected}, \cite{domingo2022optimization}, \cite{somashekarappa2023neural}, \cite{haefflinger2025datadriven}. While both lines of work can perform reliably in controlled settings, they remain limited in two aspects that are critical for unconstrained HRI: (i) many methods are tailored to fixed-configuration scenarios---e.g., specific games \cite{mishra2022knowing} or multiparty conversations with a fixed number of participants \cite{haefflinger2025datadriven}---and thus exhibit limited adaptability to unstructured real-world interactions; (ii) most approaches model gaze reasoning through one dominant class of human gaze-related regularities (e.g., mutual gaze \cite{zhang2017look} or responsive joint attention \cite{somashekarappa2023neural}). Although a few works consider multiple behavioral regularities, they often rely on simple hand-crafted scoring or prioritization strategies (e.g., computing a ``curiosity score'' with manually tuned parameters \cite{pan2020realistic}, \cite{mishra2022knowing}) to determine gaze targets, which can be less robust in complex contexts. In this work, we address these limitations with a VLM-based pipeline for anthropomorphic gaze reasoning.

\subsection{Gaze-Shift Motion Generation}

Research on gaze-shift motion generation---namely, synthesizing eye--head coordinated movements for gaze reorientation---can be roughly divided into optimization-based and learning-based approaches. Optimization-based methods encode physiology-inspired factors as energy terms, formulate an energy-minimization problem, and compute eye and head movements via numerical optimization \cite{andrist2012headeye}, \cite{liu2023control}, \cite{pan2025headeyek}. Although such methods can reproduce coordinated gaze shifts to some extent, their parametric formulations often limit the diversity and expressiveness of the generated gaze-shift kinematics. Learning-based methods instead leverage human eye--head movement datasets to train models that produce naturalistic coordinated motions \cite{le2012live}, \cite{ferstl2023generating}, \cite{goude2024realtime}. These works mainly focus on stylized gaze behaviors or broader facial animation in conversational settings, typically driven by speech content and/or explicit style controls, which can reduce their applicability beyond dialogue-centric scenarios. In contrast, we treat eye--head coordinated gaze-shift motion generation as a standalone capability and train a generative model on our self-collected human gaze-shift dataset, enabling the synthesis of diverse and human-like gaze-shift kinematics for unconstrained HRI.

\section{Methodology}

\begin{figure*}[htbp]
    \centerline{\includegraphics[width=0.9\textwidth, keepaspectratio]{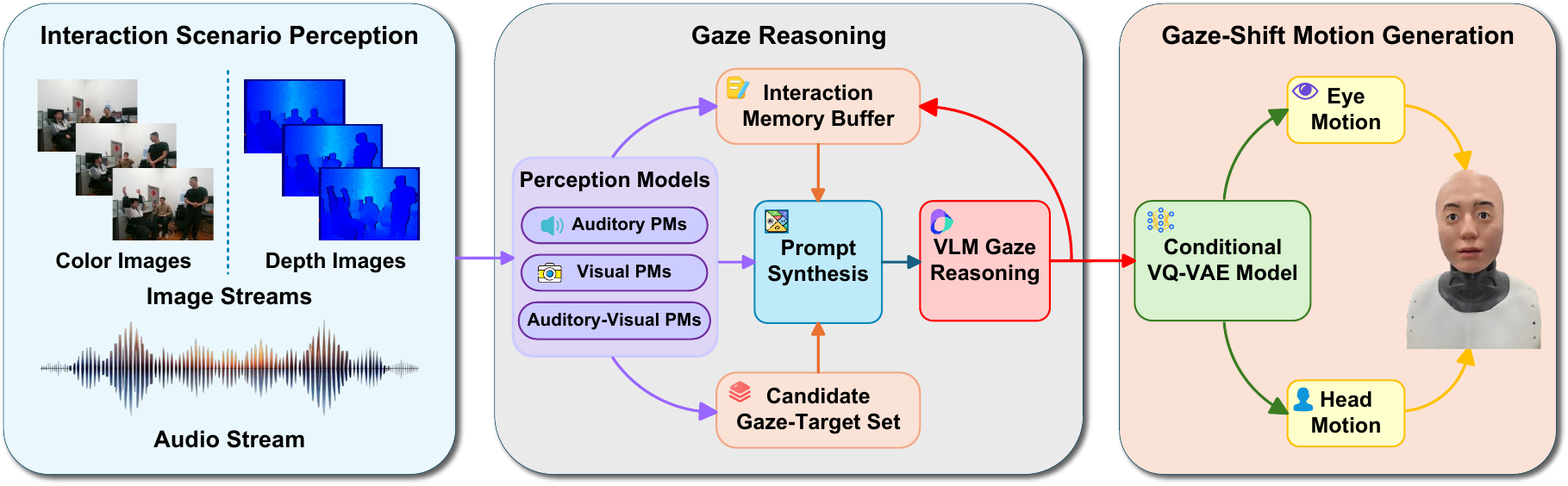}}
    \caption{Overview of our RGS framework.}
    \label{fig:rgs_overview}
\end{figure*}

This section details the proposed RGS framework and summarizes its core modules and data flow. Fig.~\ref{fig:rgs_overview} provides an overview of the framework. The framework consists of three tightly coupled stages: (i) Interaction Scenario Perception, which captures multimodal perceptual inputs from the interactive environment; (ii) Gaze Reasoning, which employs a VLM-based gaze reasoning pipeline to infer context-appropriate gaze targets; and (iii) Gaze-Shift Motion Generation, where a conditional VQ-VAE model translates the inferred targets into naturalistic gaze-shift kinematics with coordinated eye--head movements. Together, the three stages support context-aware gaze shifts in unconstrained HRI.

\subsection{Interaction Scenario Perception}

Our desktop humanoid robot head is equipped with a fixed-position RGB-D camera and a single-channel microphone. These sensors are time-synchronized to stream three modalities: a color image stream, a depth image stream, and an audio stream. The color image stream provides semantic visual cues, such as facial expressions, body pose, and scene context, which help identify candidate gaze targets. The depth image stream complements the color image stream by supplying geometric and spatial information, including 3D positions and relative distances, enabling accurate spatial localization of gaze targets. The audio stream captures realtime acoustic signals, including speech content and emotional cues, providing additional interaction information for context-appropriate gaze-target selection. Together, these modalities constitute the Interaction Scenario Perception module, producing multimodal interaction cues that serve as inputs to the subsequent Gaze Reasoning stage.

\subsection{Gaze Reasoning}

\begin{figure}[htbp]
    \centerline{\includegraphics[width=\columnwidth, keepaspectratio]{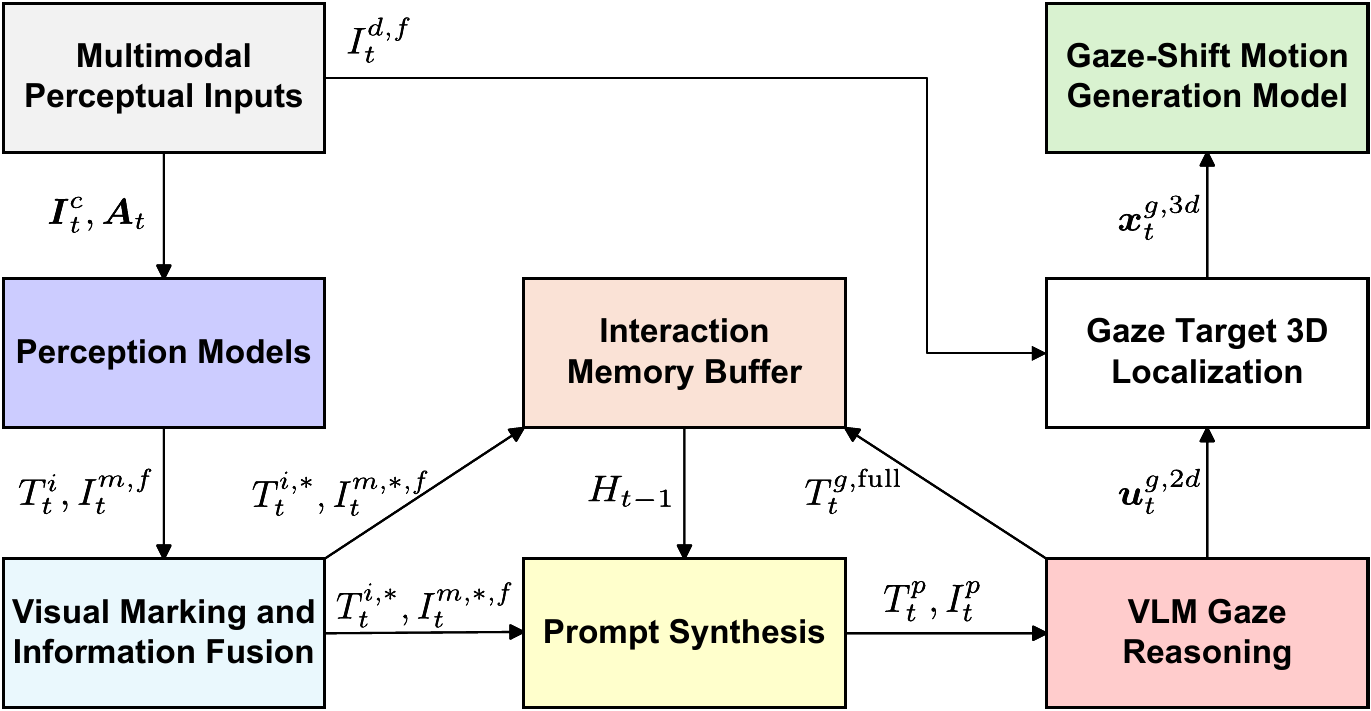}}
    \caption{VLM-Based Gaze Reasoning Pipeline.
    Notation: superscripts $f$, $m$ and $*$ denote the final frame of a cycle, instance masking, and mark indexing, respectively.}
    \label{fig:gaze_reasoning_pipeline}
\end{figure}

As illustrated in Fig.~\ref{fig:gaze_reasoning_pipeline}, the Gaze Reasoning stage estimates a context-appropriate gaze target at each inference cycle $t$ and outputs its 3D position $\boldsymbol{x}_t^{g,3d}$ in the robot base frame. The pipeline converts synchronized multimodal perception inputs into a discrete target selection and a continuous spatial estimate through the following modules.

\noindent\textbf{Multimodal Perceptual Inputs.} During cycle $t$, the robot acquires timestamp-aligned sensory streams: the color image stream $\boldsymbol{I}_t^{c}$, the depth image stream $\boldsymbol{I}_t^{d}$, and the audio stream $\boldsymbol{A}_t$. We denote by $I_t^{d,f}$ the final depth frame in this cycle, which is later used for 3D back-projection.

\noindent\textbf{Perception Models (PMs).} Given the synchronized perceptual streams $\{\boldsymbol{I}_t^{c}, \boldsymbol{A}_t\}$, a collection of pretrained auditory, visual, and auditory-visual perception models extracts high-level interaction semantics from the scene. The module produces (i) a structured textual representation $T_t^{i}$ encoding salient social and perceptual cues (e.g., tracked persons, actions, speaking status/content, and candidate objects), where the superscript $i$ denotes interaction-related information, and (ii) a masked final-frame image $I_t^{m,f}$ that spatially grounds the extracted semantics via instance-level masks.

\noindent\textbf{Visual Marking and Information Fusion.} To enable unambiguous visual grounding, we apply Set-of-Mark prompting \cite{yang2023setofmark} to assign a unique integer mark to each human/object instance in $I_t^{m,f}$. This yields a mark-annotated image $I_t^{m,*,f}$ and a mark-indexed interaction description $T_t^{i,*}$, where superscript $*$ indicates that the visual instances are explicitly indexed by marks. The set of marked instances constitutes the candidate gaze-target set in cycle $t$.

\noindent\textbf{Interaction Memory Buffer.} This buffer maintains a compact interaction memory to support temporally consistent gaze behavior. Its state is denoted by $H_{t-1}$ and includes the previous-cycle marked observations and reasoning result:
\begin{equation}
H_{t-1}=\{T_{t-1}^{i,*}, I_{t-1}^{m,*,f}, T_{t-1}^{g,\mathrm{full}}, T_{t-k:t-1}^{\mathrm{hist}}\},
\end{equation}
\noindent where $T_{t-k:t-1}^{\mathrm{hist}}$ stores longer-term textual context (e.g., dialogue over the past $k$ cycles). After inference, the buffer is updated with $\{T_t^{i,*}, I_t^{m,*,f}, T_t^{g,\mathrm{full}}\}$.

\noindent\textbf{Prompt Synthesis.} This module constructs task-specific prompts for gaze reasoning by fusing the current marked observations $\{T_t^{i,*}, I_t^{m,*,f}\}$ with the historical context $H_{t-1}$. It outputs a text prompt $T_t^{p}$ and a visual prompt $I_t^{p}$, where visual marks establish an explicit alignment between textual descriptions and image regions, and historical context supports temporally coherent reasoning.

\noindent\textbf{VLM Gaze Reasoning.} Given $\{T_t^{p}, I_t^{p}\}$, the VLM performs gaze reasoning and selects a target from the candidate set. It outputs (i) a full gaze-target record $T_t^{g,\mathrm{full}}$, including the selected mark ID, semantic category, and corresponding mask/bounding box, and (ii) the 2D gaze point $\boldsymbol{u}_t^{g,2d}$, defined as the center of the selected target's bounding box. For human targets, the face bounding-box center is used to enable natural eye contact.

\noindent\textbf{Gaze Target 3D Localization.} Finally, this module back-projects $\boldsymbol{u}_t^{g,2d}$ using the final depth frame $I_t^{d,f}$, and transforms the resulting 3D point into the robot base frame, yielding $\boldsymbol{x}_t^{g,3d}$. This 3D target position conditions the subsequent gaze-shift motion generation model.

\subsection{Gaze-Shift Motion Generation}

Given the inferred next gaze-target position $\boldsymbol{x}_t^{g,3d}\in\mathbb{R}^3$, the robot executes a gaze shift via coordinated eye--head motion. We consider a desktop humanoid robot head with limited actuation: the two eyes are modeled as coupled 2 degrees-of-freedom (DoF) rotations (yaw and pitch), and the head is modeled as a 3-DoF rotational joint (yaw, pitch, and roll) with a shared rotation center. Since gaze shifts are brief and the low-level servo control loop typically runs at a modest rate, we do not optimize high-rate trajectories. Instead, we predict a human-like \emph{motion allocation}---i.e., the eye and head rotation \emph{increments} that realize a natural gaze shift.

\begin{figure*}[htbp]
    \centerline{\includegraphics[width=0.95\textwidth, keepaspectratio]{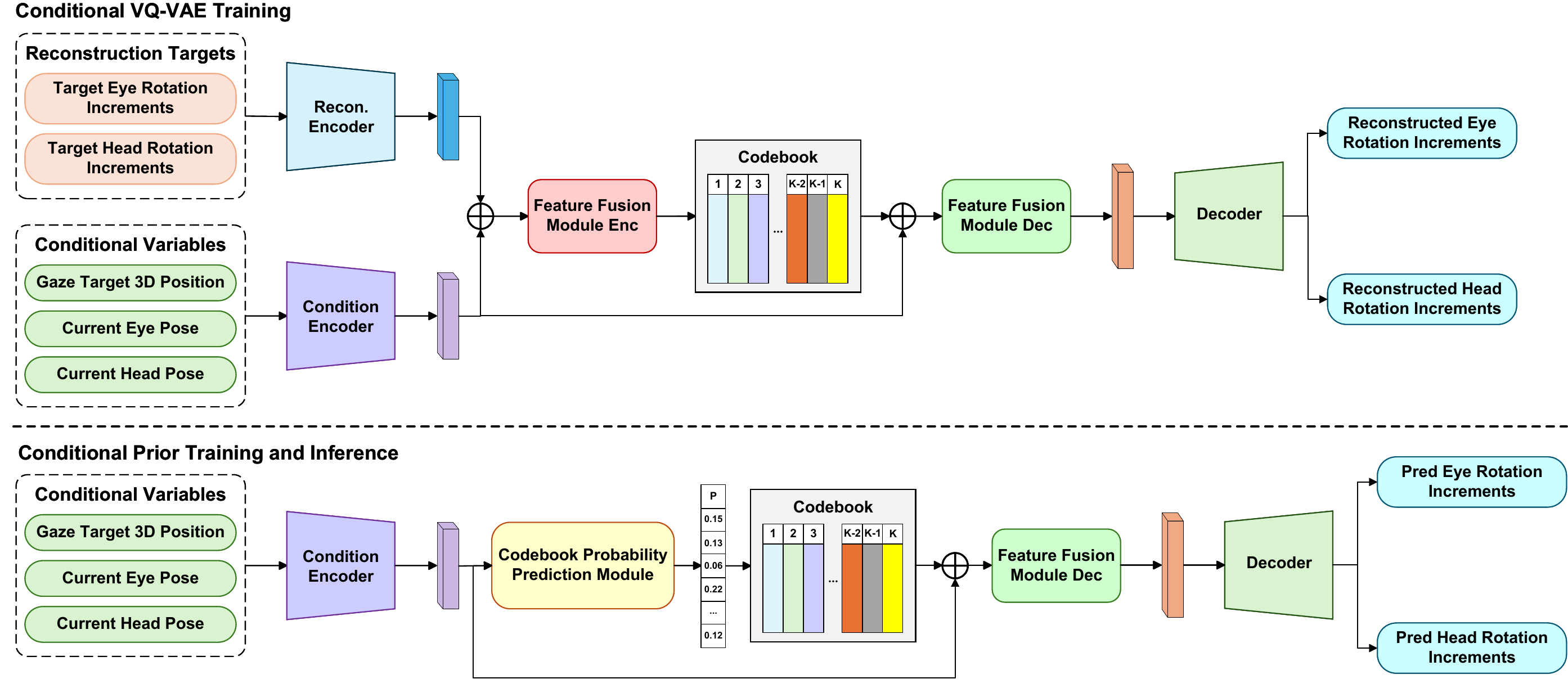}}
    \caption{Overview of the proposed gaze-shift motion generation model, trained in two stages. Top: a conditional VQ-VAE is trained to reconstruct eye--head rotation increments from ground-truth motions and conditioning inputs. Bottom: a conditional prior predicts a distribution over codebook entries from the conditioning inputs.
    The prior selects the maximum-probability code to compute the loss during training, while using stochastic sampling at inference time to enable behavioral diversity.}
    \label{fig:vqvae}
\end{figure*}

Let $\boldsymbol{\theta}^{e}\in\mathbb{R}^{2}$ and $\boldsymbol{\theta}^{h}\in\mathbb{R}^{3}$ denote the current eye and head Euler poses, respectively.
We predict
$\mathbf{y}=\{\Delta\boldsymbol{\theta}^{e},\,\Delta\boldsymbol{\theta}^{h}\}$
given the condition
$\mathbf{c}=\{\boldsymbol{\theta}^{e},\,\boldsymbol{\theta}^{h},\,\boldsymbol{x}^{g,3d}\}$.
To this end, we propose a conditional VQ-VAE trained on our human gaze-shift dataset.
Fig.~\ref{fig:vqvae} summarizes the two-stage training procedure and the inference pipeline.

\noindent\textbf{Conditional VQ-VAE Reconstruction Model.} The reconstruction model takes (i) a target motion allocation $\mathbf{y}$ (ground-truth eye/head rotation increments) and
(ii) conditional variables $\mathbf{c}$. A reconstruction encoder (Recon. Encoder in Fig.~\ref{fig:vqvae}) and a condition encoder extract features from $\mathbf{y}$ and $\mathbf{c}$, respectively. The two feature streams are fused by an encoder-side fusion module to produce a latent $\mathbf{z}_e$, which is vector-quantized by nearest-neighbor lookup in a codebook $\{\mathbf{e}_k\}_{k=1}^{K}$, yielding the quantized embedding $\mathbf{z}_q$. We then concatenate $\mathbf{z}_q$ with the condition features and apply a decoder-side fusion module, followed by a decoder to predict the reconstructed allocation $\hat{\mathbf{y}}=\{\Delta\hat{\boldsymbol{\theta}}^{e},\,\Delta\hat{\boldsymbol{\theta}}^{h}\}$.
After training, each codebook entry can be interpreted as a discrete eye--head coordination mode, while the conditional pathway enforces consistency with the current pose and target geometry.

The conditional VQ-VAE is trained with the standard objective:
\begin{equation}
\mathcal{L}_{\text{vq}}
=\mathcal{L}_{\text{rec}}
+\underbrace{\left\lVert \mathrm{sg}[\mathbf{z}_e]-\mathbf{z}_q\right\rVert_2^2}_{\mathcal{L}_{\text{embed}}}
+\beta\,\underbrace{\left\lVert \mathbf{z}_e-\mathrm{sg}[\mathbf{z}_q]\right\rVert_2^2}_{\mathcal{L}_{\text{commit}}},
\label{eq:vqvae_loss}
\end{equation}
where $\mathrm{sg}[\cdot]$ denotes the stop-gradient operator and $\beta$ is a weighting hyperparameter.

Since both eye and head motions are rotational, we use a rotation-aware geodesic reconstruction loss. Specifically, we compose the predicted Euler target poses as
\begin{equation}
\hat{\boldsymbol{\theta}}^{T,e}
=\boldsymbol{\theta}^{e}+\Delta\hat{\boldsymbol{\theta}}^{e},\qquad
\hat{\boldsymbol{\theta}}^{T,h}
=\boldsymbol{\theta}^{h}+\Delta\hat{\boldsymbol{\theta}}^{h}.
\label{eq:euler_compose}
\end{equation}
and convert them into rotation matrices
$\hat{\mathbf{R}}^{T,e}=\mathcal{R}(\hat{\boldsymbol{\theta}}^{T,e})$ and
$\hat{\mathbf{R}}^{T,h}=\mathcal{R}(\hat{\boldsymbol{\theta}}^{T,h})$.
The reconstruction loss is
\begin{equation}
\mathcal{L}_{\text{rec}}
= d_g(\hat{\mathbf{R}}^{T,e},\mathbf{R}^{T,e})+\lambda_{\text{rc}}\, d_g(\hat{\mathbf{R}}^{T,h},\mathbf{R}^{T,h}),
\label{eq:rec_geodesic}
\end{equation}
where $\mathbf{R}^{T,e}$ and $\mathbf{R}^{T,h}$ are the ground-truth target poses, and the weight $\lambda_\text{rc}$ balances the eye/head terms.
We define the geodesic distance on $\mathrm{SO}(3)$ as
\begin{equation}
d_g(\mathbf{R}_1,\mathbf{R}_2)=\cos^{-1}\!\left(\frac{\mathrm{tr}(\mathbf{R}_1\mathbf{R}_2^{\top})-1}{2}\right).
\label{eq:geodesic}
\end{equation}

\noindent\textbf{Learning a Conditional Prior over Discrete Codes.} The reconstruction model requires a code selection at inference time, but only $\mathbf{c}$ is available.
We therefore learn a conditional prior $p_\phi(z\mid\mathbf{c})$ over code indices $z\in\{1,\dots,K\}$.
For supervision, we run the trained VQ-VAE encoder on the training set and record the selected index $z^{*}$ for each sample.
The prior outputs a $K$-way probability distribution
\begin{equation}
\boldsymbol{\pi}=p_\phi(z\mid\mathbf{c})=\mathrm{Softmax}(g_\phi(\mathbf{c})),
\label{eq:prior_softmax}
\end{equation}
where $g_\phi(\cdot)$ denotes the prior network that outputs logits and $\boldsymbol{\pi}\in\mathbb{R}^{K}$.

We train the prior with (i) a focal loss for code prediction and (ii) an auxiliary motion-consistency loss that encourages the prior to select codes whose decoded motions match the ground truth. Specifically, we select the maximum-probability code $\hat z=\arg\max_k \pi_k$ and feed it into the fixed VQ-VAE decoder conditioned on $\mathbf{c}$ to obtain predicted eye/head rotation increments during training. The overall objective is:
\begin{equation}
\mathcal{L}_{\text{prior}}
=\mathcal{L}_{\text{focal}}(\boldsymbol{\pi}, z^{*})
+\eta\,\mathcal{L}_{\text{mc}}(\hat z,\mathbf{c}),
\label{eq:prior_loss}
\end{equation}
where $\eta$ balances the two loss terms.

The motion-consistency loss reuses the geodesic distance:
\begin{equation}
\mathcal{L}_{\text{mc}}
= d_g(\tilde{\mathbf{R}}^{T,e},\mathbf{R}^{T,e})+\lambda_{\text{mc}}\, d_g(\tilde{\mathbf{R}}^{T,h},\mathbf{R}^{T,h}),
\label{eq:mc}
\end{equation}
where the weight $\lambda_{\text{mc}}$ balances the two terms, and
$\tilde{\mathbf{R}}^{T,e}$ and $\tilde{\mathbf{R}}^{T,h}$ are obtained by decoding the code $\hat z$ conditioned on $\mathbf{c}$,
composing the predicted Euler target poses as in Eq.~\eqref{eq:euler_compose},
and converting the resulting poses to rotation matrices.

\noindent\textbf{Inference and Behavioral Diversity.} At runtime, given $\mathbf{c}$, the prior predicts $\boldsymbol{\pi}$.
Rather than taking maximum-probability code, we sample a code $z\sim\mathrm{Cat}(\boldsymbol{\pi})$ and decode it with $\mathbf{c}$ to obtain $\hat{\mathbf{y}}$.
This stochastic code sampling yields subtle yet plausible variations across similar interaction contexts, reducing mechanical repetition and improving perceived naturalness in unconstrained HRI.

\section{Experiments}

\subsection{Experimental Setup}

\noindent\textbf{Robot Platform and Execution.} All experiments are conducted on a desktop humanoid robot head equipped with a fixed RGB-D camera and a microphone. The full RGS framework is executed on an external industrial PC in discrete inference cycles of \SI{1.5}{\second} to balance the system responsiveness with the processing latency of upstream perception models. This interval represents the high-level reasoning rate rather than the low-level servo control frequency. The generated gaze shift is executed rapidly by the underlying servo controller, leaving the majority of the cycle for stable fixation. Crucially, this timing aligns with the natural ``saccade--fixation'' pattern of human vision \cite{rayner1998eye}: the generated gaze shift (saccade) is executed swiftly at the beginning of each cycle, while the remainder of the cycle serves as a stable fixation period, allowing the robot to dwell on the target.

\noindent\textbf{Gaze Reasoning Configuration.}
We instantiate the gaze reasoning pipeline using the vision--language model Doubao-1.5-vision-pro as a representative off-the-shelf VLM, which provides strong visual understanding with low latency, enabling efficient online inference in our \SI{1.5}{\second} cycle. For temporal consistency, the Interaction Memory Buffer maintains $k{=}10$ past cycles (\SI{15}{\second}) of context for successive inference cycles.

\noindent\textbf{Training Setup for Gaze-Shift Motion Generation Model.}
The conditional VQ-VAE is trained on a self-collected human gaze-shift dataset with 805 samples (train/val = 8:2). We use codebook size $K{=}10$, commitment weight $\beta{=}0.25$, and set $\lambda_{\text{rc}}{=}1$ and $\lambda_{\text{mc}}{=}1$ to balance the eye/head terms in Eq.~\eqref{eq:rec_geodesic} and Eq.~\eqref{eq:mc}. The conditional VQ-VAE is optimized with Adam for 200 epochs (batch size 32; learning rate $1{\times}10^{-3}$; weight decay $1{\times}10^{-4}$) using a multi-step scheduler (milestones at 100/150, decay factor 0.5). The conditional prior is trained with $\eta{=}1$ for 100 epochs using Adam (batch size 32; learning rate $1{\times}10^{-3}$; weight decay $1{\times}10^{-4}$). Checkpoints are selected on the validation set by minimizing the sum of the Mean Geodesic Distance (MGD) for eye and head rotations in each stage.

\subsection{Evaluation of Gaze Reasoning Pipeline}

We assess whether the proposed VLM-based gaze reasoning pipeline produces human-like gaze-target selection in unconstrained social scenes. The evaluation is organized around four widely observed gaze-orienting regularities reported in the psychology literature, each instantiated with scenario-based video stimuli:

\vspace{-0.3em}
\begin{itemize}\setlength{\itemsep}{0pt}\setlength{\topsep}{0pt}\setlength{\parsep}{0pt}
  \item \textbf{(H1) Deictic orienting of attention:} attention shifts toward an intended referent in response to deictic cues such as referring expressions and pointing gestures \cite{jachmann2023whena}.
  \item \textbf{(H2) Social orienting reflex:} gaze is preferentially attracted by newly appearing stimuli, with a pronounced bias toward newly entered people \cite{morand2014parietooccipitala}.
  \item \textbf{(H3) Turn-taking:} interlocutors tend to visually track the current speaker to maintain conversational coordination \cite{jokinen2013gazea}.
  \item \textbf{(H4) Responsive joint attention:} an observer aligns gaze with the entity that others are jointly attending to \cite{mundy2007attention}.
\end{itemize}
\vspace{-0.3em}

Four participants recorded 20 short clips per regularity (80 clips total) with 1--4 people and diverse everyday objects. All clips are then strictly screened to reduce confounds: four reviewers verify that (i) the intended social-attentional situation is unambiguous, and (ii) no extra salient distractors (e.g., irrelevant abrupt motion) or unexpected events (e.g., unintended entrants) are present. Only clips unanimously approved are retained for evaluation. After screening, 19 clips remained for H2 and 17 clips remained for H4.

For each retained clip, we run the pipeline online and evaluate its selected gaze target (i.e., the chosen instance from the cycle-wise candidate gaze-target set). We define $t_0$ as the earliest inference cycle where the triggering cue becomes observable (e.g., pointing onset, a new person entering, a speaker switch, or the emergence of a shared gaze direction). A trial is counted as correct if the pipeline selects the expected target in either of the next two cycles ($t_0{+}1$ or $t_0{+}2$); otherwise it is incorrect.

\begin{figure}[htbp]
  \centering
  \includegraphics[width=\columnwidth]{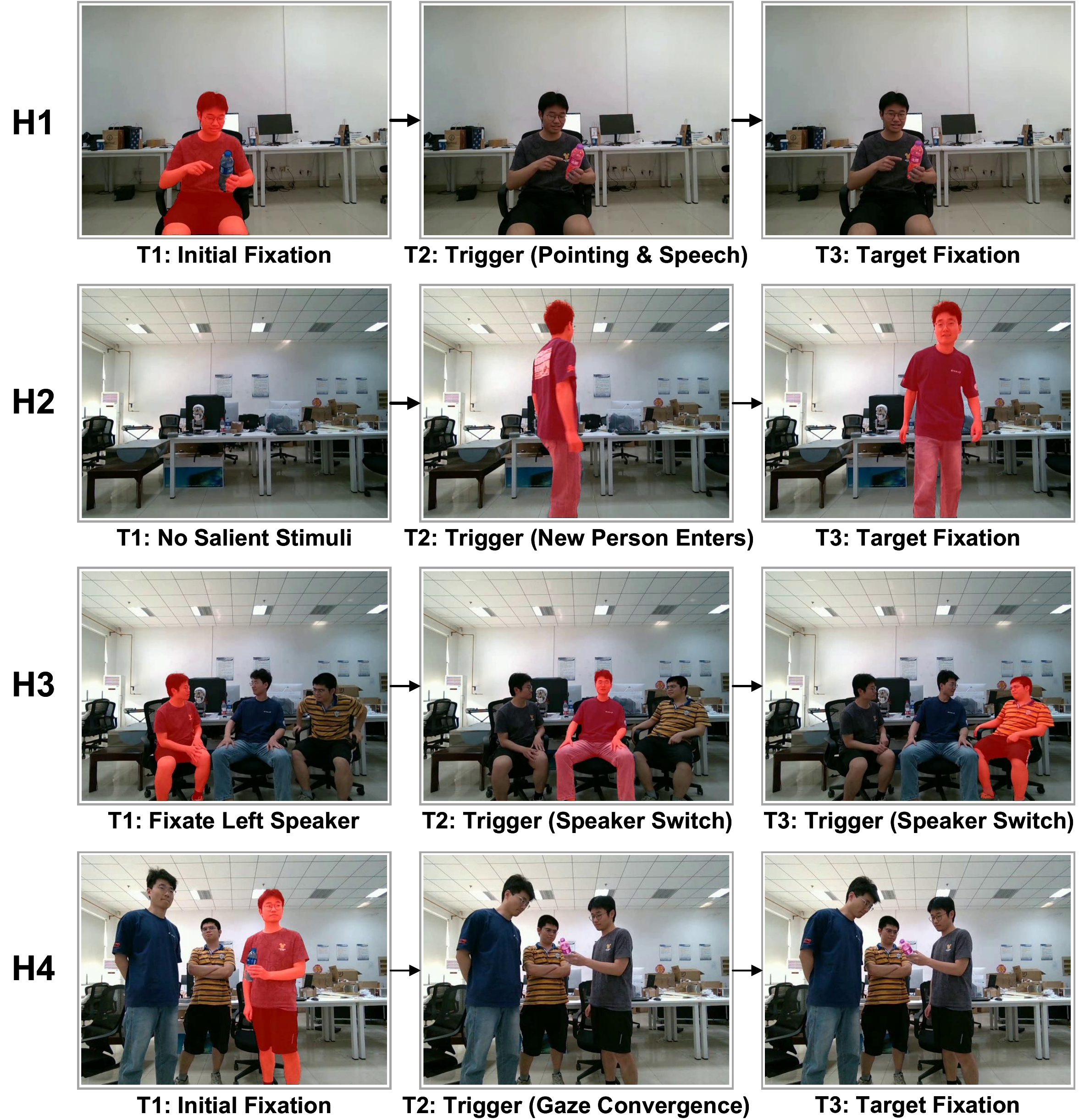}
  \caption{Qualitative examples of the proposed VLM-based gaze reasoning pipeline across four gaze-orienting regularities (H1--H4). Each row corresponds to one regularity and shows three representative inference cycles (T1--T3) selected at different times. The three cycles are temporally ordered but not necessarily consecutive. The red semi-transparent overlay indicates the gaze target selected by the pipeline at each illustrated inference cycle.}
  \label{fig:gaze_qualitative}
\end{figure}

We report qualitative examples in Fig.~\ref{fig:gaze_qualitative} and per-regularity success rates in Table~\ref{tab:gaze_reasoning_eval}. The pipeline performs strongly on H1--H3, while performance on H4 is comparatively lower yet acceptable. Error inspection suggests that several H4 failures are mainly due to missed detections of jointly attended objects under challenging configurations (e.g., partially occluded items such as an open book on a participant's lap), which prevent these objects from entering the candidate gaze-target set.

\begin{table}[htbp]
\centering
\caption{Human-likeness evaluation of gaze-target selection across four gaze-orienting regularities.}
\label{tab:gaze_reasoning_eval}
\setlength{\tabcolsep}{3.5pt}
\renewcommand{\arraystretch}{1.12}
\begin{tabular}{lccc}
\toprule
\textbf{Regularity} & \textbf{Clips} & \textbf{Correct} & \textbf{Success rate (\%)} \\
\midrule
(H1) Deictic orienting of attention  & 20 & 17 & 85.0 \\
(H2) Social orienting reflex         & 19 & 19 & 100.0 \\
(H3) Turn-taking                     & 20 & 17 & 85.0 \\
(H4) Responsive joint attention      & 17 & 12 & 70.6 \\
\bottomrule
\end{tabular}
\end{table}

\subsection{Training Analysis of Gaze-Shift Motion Generation Model}

To validate the learning effectiveness of our proposed framework, we analyze the training dynamics and convergence metrics of the two-stage training process. Specifically, we assess: (i) the reconstruction fidelity of the conditional VQ-VAE, i.e., whether the learned discrete codebook captures natural human eye--head coordination modes; (ii) the conditional prior, i.e., whether it can reliably infer discrete latent codes from the conditional variables. We report the MGD between predicted and ground-truth rotations for target eye/head poses and express it in degrees.

\begin{figure}[htbp]
  \centering

  \subfloat[Training Stage-1]{
    \includegraphics[width=0.47\linewidth]{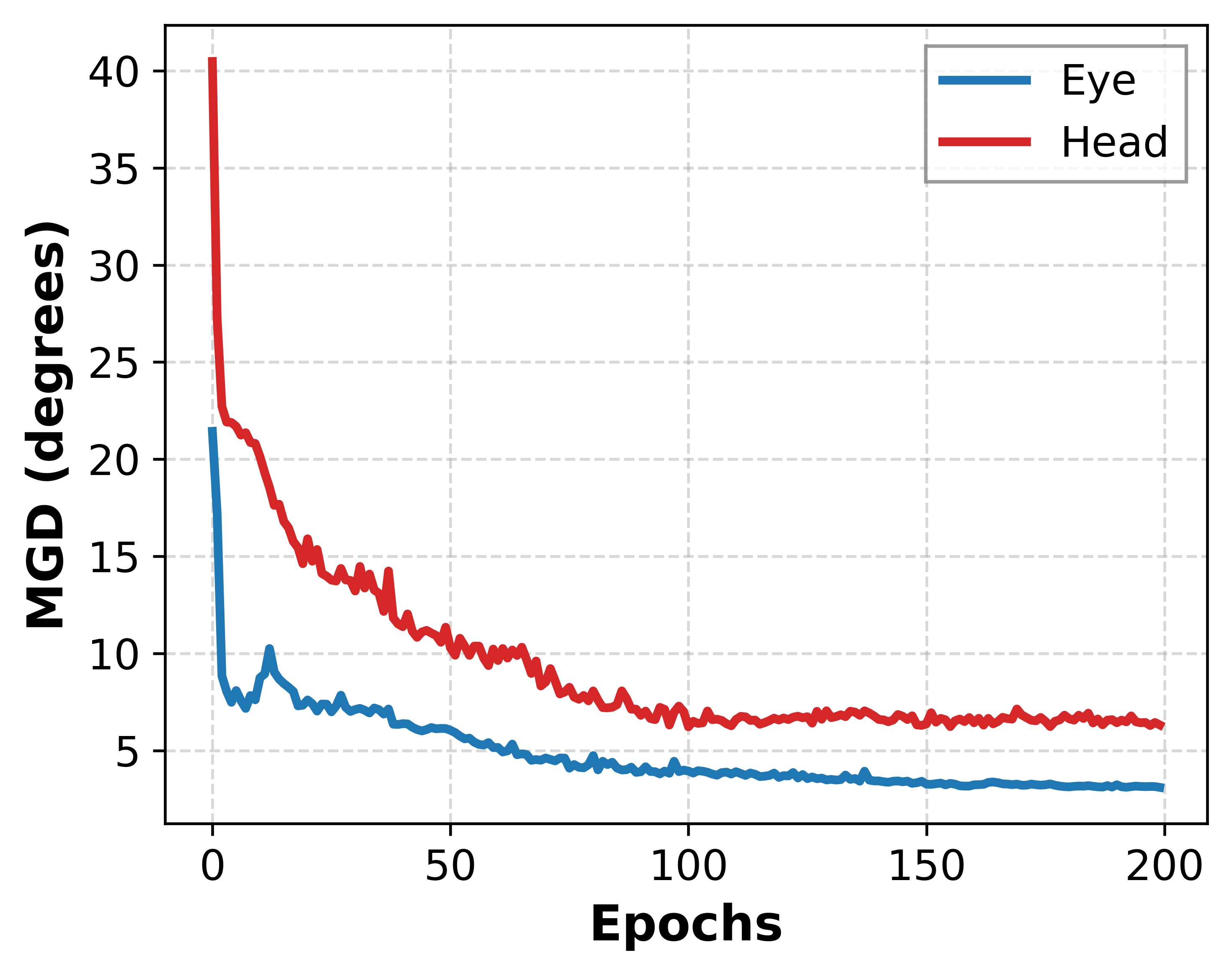}
    \label{fig:training_stage1}
  }
  \hfill
  \subfloat[Training Stage-2]{
    \includegraphics[width=0.47\linewidth]{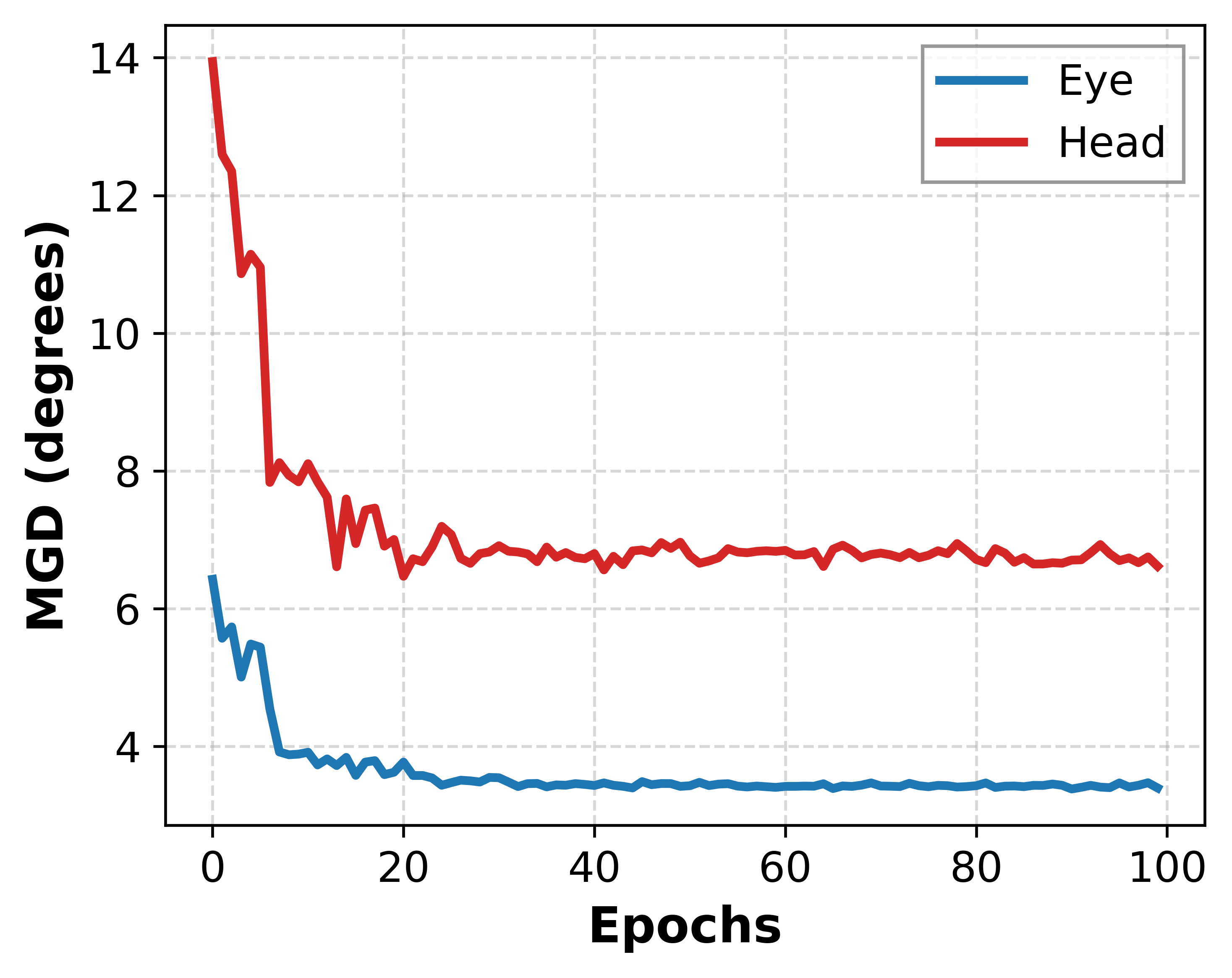}
    \label{fig:training_stage2}
  }

  \caption{
  Validation-set MGD curves for the two-stage training.
  (a) Training Stage-1: conditional VQ-VAE reconstruction errors.
  (b) Training Stage-2: conditional prior inference errors.
  Both stages exhibit consistent error reduction and stable convergence.
  }
  \label{fig:training_curves}
\end{figure}

\noindent\textbf{Training Stage-1: Conditional VQ-VAE Reconstruction.}
We first train the conditional VQ-VAE and track validation-set errors across epochs in Fig.~\ref{fig:training_curves}(a). As shown, both eye and head reconstruction errors decrease steadily and converge to stable plateaus. The best-performing model achieves an MGD of 3.1$^\circ$ for target eye pose and 6.2$^\circ$ for target head pose, indicating that the learned discrete latent representation preserves high-fidelity eye--head coordination modes.

\noindent\textbf{Training Stage-2: Conditional Prior Learning.}
We freeze the VQ-VAE codebook and decoder and train the conditional prior to predict discrete codes. As shown in Fig.~\ref{fig:training_curves}(b), the validation curves show a rapid error reduction in the early stage of training and then stabilize. The best prior achieves an MGD of 3.4$^\circ$ for target eye pose and 6.5$^\circ$ for target head pose, close to the reconstruction upper bound from Training Stage-1, suggesting accurate code inference and faithful motion synthesis under the given conditions.

\subsection{Inference-Time Diversity Evaluation of Gaze-Shift Motion Generation}

\begin{figure}[htbp]
    \centering
    \includegraphics[width=\linewidth]{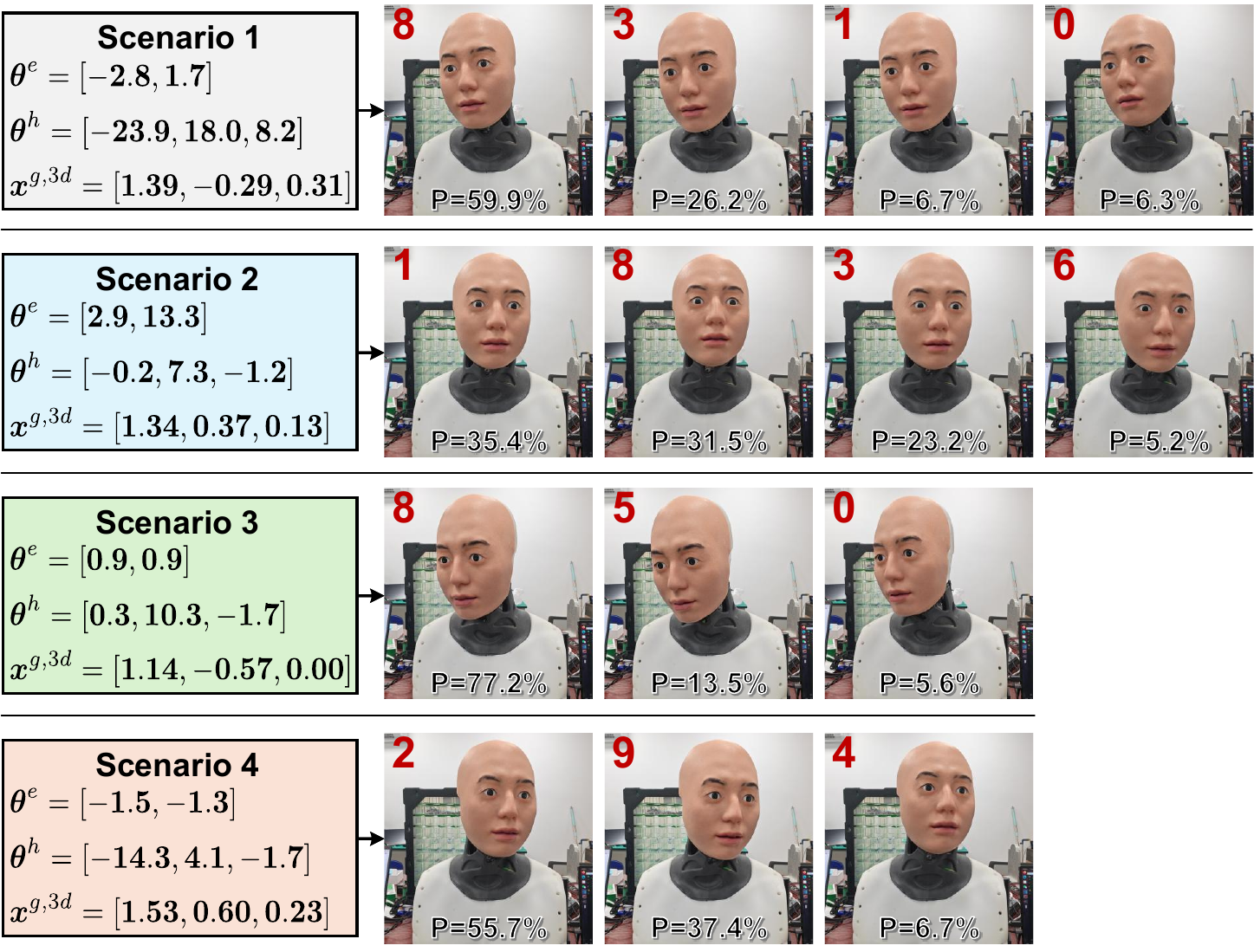}
    \caption{
    Qualitative visualization of inference-time diversity in eye--head coordinated gaze-shift motion generation.
    Given identical initial conditions $\mathbf{c}=\{\boldsymbol{\theta}^{e},\,\boldsymbol{\theta}^{h},\,\boldsymbol{x}^{g,3d}\}$, the model generates multiple distinct yet plausible eye--head coordination modes via stochastic code sampling.
    Red numbers denote sampled discrete code indices, and P indicates the corresponding prior probability (we show samples with probability $>5\%$).
    }
    \label{fig:diversity_generation}
\end{figure}

We further evaluate the generated motions on the physical robot platform to assess the naturalness and diversity of the synthesized gaze-shift behaviors in real-world settings. The experiments utilize the desktop humanoid robot head shown in Fig.~\ref{fig:diversity_generation}. A critical feature of this platform is its kinematic redundancy: the combination of 2-DoF eyes and a 3-DoF head allows the robot to reach a given gaze target through a lot of valid eye--head pose combinations. This design mimics the biological redundancy of the human oculomotor system. Consequently, whereas a deterministic regression model would likely collapse these diverse possibilities into an unnatural ``average'' motion, our generative approach is specifically designed to exploit this redundancy and produce varied, human-like behaviors.

To validate the capability of our VQ-VAE formulation to model the intrinsic stochasticity of human gaze shifts, we conduct an experiment across four interaction scenarios. In each trial, we fix the initial eye--head pose and the target gaze location, and then sample multiple codes from the learned prior to synthesize diverse eye--head motions. Fig.~\ref{fig:diversity_generation} visualizes the resulting distinct yet plausible gaze-shift behaviors (we show samples with prior probability $>5\%$). For instance, in the first scenario (top row), the model generates both a synergistic eye--head movement (Code 8) and a head-dominant shift with smaller eye rotation (Code 3) for the same target. These results confirm that our framework avoids mechanical repetition. By sampling from the learned prior, the robot exhibits diverse eye-head coordination strategies akin to human behavioral variability, thereby enhancing the perceived naturalness in unconstrained HRI.

\section{Conclusion}

This paper introduced the RGS framework to humanize robot gaze shifts in unconstrained HRI by coupling (i) a VLM-based gaze reasoning pipeline that infers context-appropriate gaze targets and (ii) a conditional VQ-VAE gaze-shift motion generation model that synthesizes naturalistic and diverse eye--head coordinated movements. Experimental results demonstrate that the gaze reasoning pipeline selects targets consistent with human gaze-orienting regularities, while the motion generation model captures characteristic eye--head coordination modes and realistic variability. A key limitation is that the two modules were evaluated largely in isolation; future work will tightly integrate them, validate end-to-end performance with user studies, and further extend the system from reactive gaze responses toward intent-driven, proactive gaze behaviors.

\bibliographystyle{IEEEtran}
\bibliography{Reference}

@article{admoni2017social,
  author = {Admoni, Henny and Scassellati, Brian},
  title = {Social Eye Gaze in Human-Robot Interaction: A Review},
  journal = {Journal of Human-Robot Interaction},
  volume = {6},
  number = {1},
  pages = {25},
  year = {2017},
  issn = {2163-0364},
  shorttitle = {Social Eye Gaze in Human-Robot Interaction},
  doi = {10.5898/JHRI.6.1.Admoni},
  langid = {english}
}

@article{vsabanovic2023robots,
  title={“Robots for good”: Ten defining questions},
  author={{\v{S}}abanovi{\'c}, Selma and Charisi, Vicky and Belpaeme, Tony and Bethel, Cindy L and Matari{\'c}, Maja and Murphy, Robin and Levy-Tzedek, Shelly},
  journal={Science Robotics},
  volume={8},
  number={84},
  pages={eadl4238},
  year={2023},
  publisher={American Association for the Advancement of Science}
}

@article{liu2024unlocking,
  author = {Liu, Xiaofeng and Ni, Rongrong and Yang, Biao and Song, Siyang and Cangelosi, Angelo},
  title = {Unlocking Human-Like Facial Expressions in Humanoid Robots: A Novel Approach for Action Unit Driven Facial Expression Disentangled Synthesis},
  journal = {IEEE Transactions on Robotics},
  volume = {40},
  pages = {3850--3865},
  year = {2024},
  issn = {1941-0468},
  shorttitle = {Unlocking Human-Like Facial Expressions in Humanoid Robots},
  doi = {10.1109/TRO.2024.3422051}
}

@inproceedings{jokinen2025domain,
  author = {Jokinen, Kristiina and Wilcock, Graham},
  title = {Towards Domain Graphs and Dialogue Graphs for Conversational Grounding in HRI},
  pages = {1373--1377},
  year = {2025},
  booktitle = {2025 20th ACM/IEEE International Conference on Human-Robot Interaction (HRI)},
  doi = {10.1109/HRI61500.2025.10973841}
}

@article{he2024omnih2o,
  author = {He, Tairan and Luo, Zhengyi and He, Xialin and Xiao, Wenli and Zhang, Chong},
  title = {OmniH2O: Universal and Dexterous Human-to-Humanoid Whole-Body Teleoperation and Learning},
  journal = {arXiv preprint arXiv:2406.08858},
  year = {2024},
  shorttitle = {OmniH2O},
  doi = {10.48550/arXiv.2406.08858},
  langid = {english},
  pubstate = {prepublished}
}

@article{zhou2023orienting,
  author = {Zhou, Ji and Hormigo, Sebastian and Busel, Natan and Castro-Alamancos, Manuel A.},
  title = {The Orienting Reflex Reveals Behavioral States Set by Demanding Contexts: Role of the Superior Colliculus},
  journal = {The Journal of Neuroscience},
  volume = {43},
  number = {10},
  pages = {1778--1796},
  year = {2023},
  issn = {0270-6474, 1529-2401},
  shorttitle = {The Orienting Reflex Reveals Behavioral States Set by Demanding Contexts},
  doi = {10.1523/JNEUROSCI.1643-22.2023},
  langid = {english}
}

@article{mundy2007attention,
  author = {Mundy, Peter and Newell, Lisa},
  title = {Attention, Joint Attention, and Social Cognition},
  journal = {Current Directions in Psychological Science},
  volume = {16},
  number = {5},
  pages = {269--274},
  year = {2007},
  issn = {0963-7214, 1467-8721},
  doi = {10.1111/j.1467-8721.2007.00518.x},
  langid = {english}
}

@inproceedings{mishra2022knowing,
  author = {Mishra, Chinmaya and Skantze, Gabriel},
  title = {Knowing Where to Look: A Planning-based Architecture to Automate the Gaze Behavior of Social Robots},
  pages = {1201--1208},
  year = {2022},
  shorttitle = {Knowing Where to Look},
  booktitle = {2022 31st IEEE International Conference on Robot and Human Interactive Communication (RO-MAN)},
  doi = {10.1109/RO-MAN53752.2022.9900740},
  langid = {english}
}

@article{haefflinger2025datadriven,
  author = {Haefflinger, Léa and Elisei, Frédéric and Bailly, Gérard},
  title = {Data-Driven Control of Eye and Head Movements for Triadic Human-Robot Interactions},
  journal = {International Journal of Social Robotics},
  volume = {17},
  number = {6},
  pages = {1075--1096},
  year = {2025},
  issn = {1875-4791, 1875-4805},
  doi = {10.1007/s12369-025-01245-2},
  langid = {english}
}

@inproceedings{zhang2017look,
  title={Look but don’t stare: Mutual gaze interaction in social robots},
  author={Zhang, Yanxia and Beskow, Jonas and Kjellstr{\"o}m, Hedvig},
  booktitle={International conference on social robotics},
  pages={556--566},
  year={2017},
  organization={Springer}
}

@article{correia2023robotic,
  author = {Correia, Filipa and Campos, Joana and Melo, Francisco S. and Paiva, Ana},
  title = {Robotic Gaze Responsiveness in Multiparty Teamwork},
  journal = {International Journal of Social Robotics},
  volume = {15},
  number = {1},
  pages = {27--36},
  year = {2023},
  issn = {1875-4791, 1875-4805},
  doi = {10.1007/s12369-022-00955-1},
  langid = {english}
}

@inproceedings{pan2020realistic,
  author = {Pan, Matthew K.X.J. and Choi, Sungjoon and Kennedy, James and McIntosh, Kyna and Zamora, Daniel Campos},
  title = {Realistic and Interactive Robot Gaze},
  pages = {11072--11078},
  year = {2020},
  issn = {2153-0866},
  booktitle = {2020 IEEE/RSJ International Conference on Intelligent Robots and Systems (IROS)},
  doi = {10.1109/IROS45743.2020.9341297}
}

@inproceedings{vogel2008targetdirected,
  author = {Vogel, Julia and de Freitas, Nando},
  title = {Target-directed attention: Sequential decision-making for gaze planning},
  pages = {2372--2379},
  year = {2008},
  issn = {1050-4729},
  shorttitle = {Target-directed attention},
  booktitle = {2008 IEEE International Conference on Robotics and Automation},
  doi = {10.1109/ROBOT.2008.4543568}
}

@article{domingo2022optimization,
  author = {Domingo, Jaime Duque and Gómez-García-Bermejo, Jaime and Zalama, Eduardo},
  title = {Optimization and improvement of a robotics gaze control system using LSTM networks},
  journal = {Multimedia Tools and Applications},
  volume = {81},
  number = {3},
  pages = {3351--3368},
  year = {2022},
  issn = {1380-7501, 1573-7721},
  doi = {10.1007/s11042-021-11112-7},
  langid = {english}
}

@inproceedings{somashekarappa2023neural,
  author = {Somashekarappa, Vidya and Sayeed, Asad and Howes, Christine},
  title = {Neural Network Implementation of Gaze-Target Prediction for Human-Robot Interaction},
  pages = {2238--2244},
  year = {2023},
  issn = {1944-9437},
  booktitle = {2023 32nd IEEE International Conference on Robot and Human Interactive Communication (RO-MAN)},
  doi = {10.1109/RO-MAN57019.2023.10309483}
}

@article{pan2025headeyek,
  author = {Pan, Yifang and Sidenmark, Ludwig and Singh, Karan},
  title = {Head-EyeK: Head-Eye Coordination and Control Learned in Virtual Reality},
  journal = {IEEE Transactions on Visualization and Computer Graphics},
  volume = {31},
  number = {10},
  pages = {9039--9051},
  year = {2025},
  issn = {1077-2626, 1941-0506, 2160-9306},
  shorttitle = {Head-EyeK},
  doi = {10.1109/TVCG.2025.3589333},
  langid = {english}
}

@article{liu2023control,
  author = {Liu, Xiaorui and Jiang, Wanyue and Su, Hang and Qi, Wen and Ge, Shuzhi Sam},
  title = {A Control Strategy of Robot Eye-Head Coordinated Gaze Behavior Achieved for Minimized Neural Transmission Noise},
  journal = {IEEE/ASME Transactions on Mechatronics},
  volume = {28},
  number = {2},
  pages = {956--966},
  year = {2023},
  issn = {1083-4435, 1941-014X},
  doi = {10.1109/TMECH.2022.3210592},
  langid = {english}
}

@inproceedings{andrist2012headeye,
  title={A head-eye coordination model for animating gaze shifts of virtual characters},
  author={Andrist, Sean and Pejsa, Tomislav and Mutlu, Bilge and Gleicher, Michael},
  booktitle={Proceedings of the 4th Workshop on Eye Gaze in Intelligent Human Machine Interaction},
  pages={1--6},
  year={2012}
}

@article{le2012live,
  title={Live speech driven head-and-eye motion generators},
  author={Le, Binh H and Ma, Xiaohan and Deng, Zhigang},
  journal={IEEE Transactions on Visualization and Computer Graphics},
  volume={18},
  number={11},
  pages={1902--1914},
  year={2012},
  publisher={IEEE}
}

@inproceedings{ferstl2023generating,
  author = {Ferstl, Ylva},
  title = {Generating Emotionally Expressive Look-At Animation},
  pages = {1--6},
  year = {2023},
  publisher = {ACM},
  location = {Rennes France},
  doi = {10.1145/3623264.3624438},
  isbn = {979-8-4007-0393-5},
  langid = {english},
  booktitle = {MIG '23: The 16th ACM SIGGRAPH Conference on Motion, Interaction and Games}
}

@article{goude2024realtime,
  author = {Goudé, Ific and Bruckert, Alexandre and Olivier, Anne-Hélène and Pettré, Julien and Cozot, Rémi},
  title = {Real-Time Multi-Map Saliency-Driven Gaze Behavior for Non-Conversational Characters},
  journal = {IEEE Transactions on Visualization and Computer Graphics},
  volume = {30},
  number = {7},
  pages = {3871--3883},
  year = {2024},
  issn = {1077-2626, 1941-0506, 2160-9306},
  doi = {10.1109/TVCG.2023.3244679},
  langid = {english}
}

@article{bristow2007social,
  author = {Bristow, Davina and Rees, Geraint and Frith, Christopher D.},
  title = {Social interaction modifies neural response to gaze shifts},
  journal = {Social Cognitive and Affective Neuroscience},
  volume = {2},
  number = {1},
  pages = {52--61},
  year = {2007},
  issn = {1749-5024, 1749-5016},
  doi = {10.1093/scan/nsl036},
  langid = {english}
}

@article{frischen2007gaze,
  author = {Frischen, Alexandra and Bayliss, Andrew P. and Tipper, Steven P.},
  title = {Gaze cueing of attention: Visual attention, social cognition, and individual differences},
  journal = {Psychological Bulletin},
  volume = {133},
  number = {4},
  pages = {694--724},
  year = {2007},
  issn = {1939-1455, 0033-2909},
  shorttitle = {Gaze cueing of attention},
  doi = {10.1037/0033-2909.133.4.694},
  langid = {english}
}

@article{yang2023setofmark,
  author = {Yang, Jianwei and Zhang, Hao and Li, Feng and Zou, Xueyan and Li, Chunyuan and Gao, Jianfeng},
  title = {Set-of-Mark Prompting Unleashes Extraordinary Visual Grounding in GPT-4V},
  journal = {arXiv preprint arXiv:2310.11441},
  year = {2023},
  doi = {10.48550/arXiv.2310.11441},
  pubstate = {prepublished}
}

@article{jachmann2023whena,
  author = {Jachmann, Torsten Kai and Drenhaus, Heiner and Staudte, Maria and Crocker, Matthew W.},
  title = {When a look is enough: Neurophysiological correlates of referential speaker gaze in situated comprehension},
  journal = {Cognition},
  volume = {236},
  pages = {105449},
  year = {2023},
  issn = {00100277},
  shorttitle = {When a look is enough},
  doi = {10.1016/j.cognition.2023.105449},
  langid = {english}
}

@article{morand2014parietooccipitala,
  author = {Morand, S. M. and Harvey, M. and Grosbras, M.-H.},
  title = {Parieto-Occipital Cortex Shows Early Target Selection to Faces in a Reflexive Orienting Task},
  journal = {Cerebral Cortex},
  volume = {24},
  number = {4},
  pages = {898--907},
  year = {2014},
  issn = {1047-3211, 1460-2199},
  doi = {10.1093/cercor/bhs368},
  langid = {english}
}

@article{jokinen2013gazea,
  author = {Jokinen, Kristiina and Furukawa, Hirohisa and Nishida, Masafumi and Yamamoto, Seiichi},
  title = {Gaze and turn-taking behavior in casual conversational interactions},
  journal = {ACM Transactions on Interactive Intelligent Systems},
  volume = {3},
  number = {2},
  pages = {1--30},
  year = {2013},
  issn = {2160-6455, 2160-6463},
  doi = {10.1145/2499474.2499481},
  langid = {english}
}

@article{rayner1998eye,
  title={Eye Movements in Reading and Information Processing: 20 Years of Research},
  author={Rayner, Keith},
  journal={Psychological Bulletin},
  volume={124},
  number={3},
  pages={372},
  year={1998},
  publisher={American Psychological Association}
}

\end{document}